\documentclass[11pt,a4paper]{article}
\usepackage[hyperref]{acl2021}
\usepackage{times}
\usepackage{latexsym}

\usepackage{microtype}

\aclfinalcopy 
\def\aclpaperid{***} 

\newcommand\BibTeX{B\textsc{ib}\TeX}

\usepackage{multirow}
\usepackage{graphicx}
\usepackage{float}
\usepackage{subcaption}
\graphicspath{ {./img/} }

\usepackage{verbatim}
\usepackage[frozencache=true,cachedir=minted-cache]{minted} 
\usepackage{nert}

\usepackage{enumitem}
\usepackage{listings}

\newcommand{\mtdata}{\textsc{MTData}}
\newcommand{\nlcodec}{\textsc{NLCodec}}
\newcommand{\nldb}{\textsc{NLDb}}
\newcommand{\rtg}{\textsc{RTG}}

\newcommand{\sentpiece}{SentencePiece}
\newcommand{\sacrebleu}{\textsc{SacreBleu}}

\newcommand{\hftok}{HuggingfaceTokenizers}

\title{Many-to-English Machine Translation Tools, Data, and Pretrained Models }

\author{Thamme Gowda \\
  Information Sciences Institute \\
  University of Southern California \\
  \eml{tg@isi.edu} \\
\And
  Zhao Zhang \\
 Texas Advanced Computing Center \\
  University of Texas \\
  \eml{zzhang@tacc.utexas.edu} \\
 \AND 
 Chris A Mattmann \\
  NASA Jet Propulsion Laboratory \\ 
  California Institute of Technology \\
  \eml{chris.a.mattmann@jpl.nasa.gov}\\
 \And
  Jonathan May \\
  Information Sciences Institute \\
  University of Southern California \\
 \eml{jonmay@isi.edu} \\
  }

\date{}

\begin{document}
\maketitle
\begin{abstract}
While there are more than 7000 languages in the world, most translation research efforts have targeted a few high resource languages.
Commercial translation systems support only one hundred languages or fewer, and do not make these models available for transfer to low resource languages. 
In this work, we present useful tools for machine translation research: \mtdata, \nlcodec, and \rtg.
We demonstrate their usefulness by creating a multilingual neural machine translation model capable of translating from 500 source languages to English. 
We make this multilingual model readily downloadable and usable as a service, or as a parent model for transfer-learning to even lower-resource languages.\footnote{Demo website: \url{http://rtg.isi.edu/many-eng}.\\ Video demo: \url{https://youtu.be/NSY0-MvO1KE}.}

\end{abstract}

\section{Introduction}

Neural machine translation (NMT)~\cite{bahdanau2014nmtattn,vaswani2017attention} has progressed to reach human performance on select benchmark tasks~\cite{barrault-etal-2019-findings,barrault-etal-2020-findings}. 
However, as MT research has mainly focused on translation between a small number of high resource languages, the unavailability of usable-quality translation models for low resource languages remains an ongoing concern.
Even those commercial translation services attempting to broaden their language coverage has only reached around one hundred languages; this excludes most of the thousands of languages used around the world today.

Freely available corpora of parallel data for many languages are available, though they are hosted at various sites, and are in various forms. A challenge for incorporating more languages into MT models is a lack of easy access to all of these datasets.While standards like ISO 639-3 have been established to bring consistency to the labeling of language resources, these are not yet widely adopted.
In addition, scaling experimentation to several hundred languages on large corpora involves a significant engineering effort.
Simple tasks such as dataset preparation, vocabulary creation, transformation of sentences into sequences, and training data selection becomes formidable at scale due to corpus size and heterogeneity of data sources and file formats.
We have developed tools to precisely address all these challenges, which we demonstrate in this work.
 
Specifically, we offer three tools which can be used either independently or in combination to advance NMT research on a wider set of languages (Section \ref{sec:tools}): firstly, \mtdata, which helps to easily obtain parallel datasets (Section \ref{sec:mtdata}); secondly, \nlcodec, a vocabulary manager and storage layer for transforming sentences to integer sequences, that is efficient and scalable (Section \ref{sec:nlcodec}); and lastly, \rtg, a feature-rich Pytorch-backed NMT toolkit that supports reproducible experiments (Section \ref{sec:rtg}).

We demonstrate the capabilities of our tools by preparing a massive bitext dataset with more than 9 billion tokens per side, and training a single multilingual NMT model capable of translating 500 source languages to English (Section \ref{sec:500-eng}).
We show that the multilingual model is usable either as a service for translating several hundred languages to English (Section \ref{sec:value.off-shelf-mt}), or as a parent model in a transfer learning setting for improving translation of low resource languages (Section \ref{sec:value.transfer-learning}). 

\section{Tools}
\label{sec:tools}
 Our tools are organized into the following sections:

\subsection{\textsc{MTData}}
\label{sec:mtdata}
\textsc{MTData} addresses an important yet often overlooked challenge -- dataset preparation. 
By assigning an ID for datasets, we establish a clear way of communicating the exact datasets used for MT experiments, which helps in reproducing the experimental setup.
By offering a unified interface to datasets from many heterogeneous sources, \mtdata\ hides mundane tasks such as locating URLs, downloading, decompression, parsing, and sanity checking. Some noteworthy features are:
\begin{itemize}[noitemsep,topsep=0pt,leftmargin=4mm]
\item \textit{Indexer}: a large index of publicly available parallel datasets.
\item \textit{Normalizer:} maps language names to ISO-639-3 codes which has representation space for 7800+ languages.\footnote{\url{https://iso639-3.sil.org}}
\item \textit{Parsers:} parses heterogeneous data formats for parallel datasets, and produces a simple plain text file by merging all the selected datasets.
\item \textit{Extensible:} new datasets and parsers can be easily added.
\item \textit{Local Cache}: reduces network transfers by maintaining a local cache, which is shared between experiments.
\item \textit{Sanity Checker}: performs basic sanity checks such as segment count matching and empty segment removal. When error states are detected, stops the setup with useful error messages.
\item \textit{Reproducible:} stores a signature file that can be used to recreate the dataset at a later time.
\item \textit{Courtesy:} shows the original \BibTeX\ citation attributed to datasets.
\item \textit{Easy Setup:} \texttt{pip install mtdata}
\item \textit{Open-source:} \\ \url{https://github.com/thammegowda/mtdata}
\end{itemize}

Listing \ref{lst:mtdata-eg} shows an example for listing and getting datasets for German-English.
\begin{listing}
\begin{minted}[baselinestretch=1, fontsize=\footnotesize, frame=lines, framesep=2mm,]{bash}
# List all the available datasets for deu-eng
$ mtdata list -l deu-eng
# Get the selected training & held-out sets
$ mtdata get -l deu-eng --merge\
 -tr wmt13_europarl_v7 wmt13_commoncrawl\ 
    wmt18_news_commentary_v13\
 -ts newstest201{8,9}_deen -o data
\end{minted}
\caption{\mtdata\ examples for listing and downloading German-English datasets.
The \texttt{--merge} flag results in merging all of the training datasets specified by \texttt{-tr} argument into a single file. }
\label{lst:mtdata-eg}
\end{listing}
 In Section~\ref{sec:datasets}, we use \mtdata\footnote{At the time of writing, v0.2.8} to obtain thousands of publicly available datasets for a large many-to-English translation experiment.

\subsection{\nlcodec}
\label{sec:nlcodec}
\nlcodec\ is a vocabulary manager with encoding-decoding schemes to transform natural language sentences to and from integer sequences.\\
\textbf{Features:}
\begin{itemize}[noitemsep,topsep=0pt,leftmargin=4mm]
\item \textit{Versatile:} Supports commonly used vocabulary schemes such as characters, words, and byte-pair-encoding (BPE) subwords~\cite{sennrich-etal-2016-BPE}.
\item \textit{Scalable:} Apache Spark\footnote{\url{https://spark.apache.org/}}\cite{zaharia2016spark} backend can be optionally used to create vocabulary from massive datasets.
\item \textit{Easy Setup:} \texttt{pip install nlcodec} 
\item \textit{Open-source:}\\ \url{https://github.com/isi-nlp/nlcodec/}
\end{itemize}

When the training datasets are too big to be kept in the primary random access memory (RAM), the use of secondary storage is inevitable.
The training processes requiring random examples lead to random access from a secondary storage device.
Even though the latest advancements in secondary storage technology such as solid-state drive (SSD) have faster serial reads and writes, their random access speeds are significantly lower than that of RAM. 
To address these problems, we include an efficient storage and retrieval layer, \nldb, which has the following features:
\begin{itemize}[noitemsep,topsep=0pt,leftmargin=4mm]
  \item \textit{Memory efficient} by adapting datatypes based on vocabulary size. For instance, encoding with vocabulary size less than 256 (such as characters) can be efficiently represented using 1-byte unsigned integers. Vocabularies with fewer than 65,536 types, such as might be generated when using subword models \cite{sennrich-etal-2016-BPE} require only 2-byte unsigned integers, and 4-byte unsigned integers are sufficient for vocabularies up to 4 billion types. 
As the default implementation of Python, CPython, uses 28 bytes for all integers, we accomplish this using NumPy~\cite{harris2020numpy}. This optimization makes it possible to hold a large chunk of training data in smaller RAM, enabling a fast random access.
  \item \textit{Parallelizable:} Offers a multi-part database by horizontal sharding that supports parallel writes (e.g., Apache Spark) and parallel reads (e.g., distributed training).
  \item Supports commonly used batching mechanisms such as random batches with approximately-equal-length sequences.
\end{itemize}

\nldb\ has a minimal footprint and is part of the \nlcodec\ package.
In Section~\ref{sec:500-eng}, we take advantage of the scalability and efficiency aspects of \nlcodec\ and \nldb\ to process a large parallel dataset with 9 billion tokens on each side.

\subsection{\rtg}
\label{sec:rtg}
Reader translator generator (\rtg) is a neural machine translation (NMT) toolkit based on Pytorch~\cite{NEURIPS2019_Pytorch}. 
Notable features of \rtg\ are:
\begin{itemize}[noitemsep,topsep=0pt,leftmargin=4mm]
\item \textit{Reproducible:} All the required parameters of an experiment are included in a single YAML configuration file, which can be easily stored in a version control system such as \texttt{git} or shared with collaborators.
\item Implements Transformer~\cite{vaswani2017attention}, and recurrent neural networks (RNN) with cross-attention models ~\cite{bahdanau2014nmtattn,luong2015effectiveAttn}.
\item Supports distributed training on multi-node multi-GPUs, gradient accumulation, and Float16 operations.
\item Integrated Tensorboard helps in visualizing training and validation losses. 
\item Supports weight sharing~\cite{press-wolf-2017-embeddings}, parent-child transfer~\cite{zoph-etal-2016-transfer}, beam decoding with length normalization~\cite{wu-etal-2016-GNMT}, early stopping, and checkpoint averaging. 
\item Flexible vocabulary options with \nlcodec\ and \sentpiece~\cite{kudo-richardson-2018-sentencepiece} which can be either shared or separated between source and target languages.
\item \textit{Easy setup:} \texttt{pip install rtg}
\item \textit{Open-source:} \url{https://isi-nlp.github.io/rtg/}
\end{itemize}

\section{Many-to-English Multilingual NMT}
\label{sec:500-eng}
In this section, we demonstrate the use of our tools by creating a massively multilingual NMT model from publicly available datasets. 

\subsection{Dataset}
\label{sec:datasets}
We use \mtdata\ to download datasets from various sources, given in Table~\ref{tab:data-sources}. 
To minimize data imbalance, we select only a subset of the datasets available for high resource languages, and select all available datasets for low resource languages. The selection is aimed to increase the diversity of data domains and quality of alignments. 

 \begin{table}[ht]
 \centering
 \footnotesize
 \begin{tabular}{p{0.3\linewidth} p{0.6\linewidth}}
  Dataset   & Reference \\ \hline\hline
 Europarl    & \citet{koehn2005europarl} \\
 KFTT Ja-En & \citet{neubig11kftt}  \\ 
 Indian6     & \citet{post-etal-2012-constructing}   \\ 
 OPUS        & \citet{tiedemann-2012-parallel}  \\
 UNPCv1     & \citet{ziemski-etal-2016-unpc}   \\
 Tilde MODEL & \citet{rozis-skadins-2017-tilde}  \\
 TEDTalks    & \citet{qi-etal-2018-pretrainemb}  \\ 
 IITB Hi-En & \citet{kunchukuttan-etal-2018-iit} \\
 Paracrawl   & \citet{espla-etal-2019-paracrawl} \\
 WikiMatrix & \citet{schwenk-etal-2019-wikimatrixv1} \\
 JW300       & \citet{agic-vulic-2019-jw300}  \\
 PMIndia & \citet{haddow2020pmindia}  \\
 OPUS100    & \citet{zhang-etal-2020-multiling-nmt} \\
 WMT [13-20] & \citet{bojar-etal-2013-findings, bojar-etal-2014-findings, bojar-etal-2015-findings, bojar-etal-2016-findings, bojar-etal-2017-findings, bojar-etal-2018-findings, barrault-etal-2019-findings, barrault-etal-2020-findings} \\

 \end{tabular}
 \caption{Various sources of MT datasets.}
 \label{tab:data-sources}
\end{table}

\begin{figure*}[ht]
\centering
    \includegraphics[width=\linewidth]{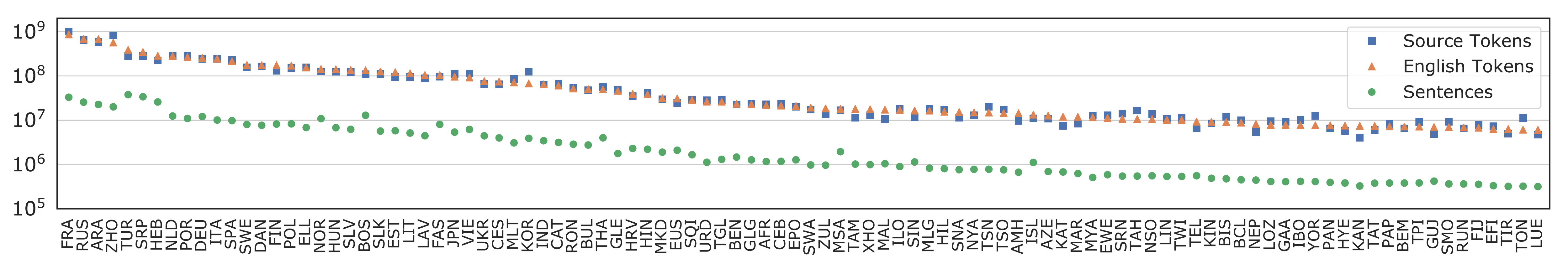}
    \includegraphics[width=\linewidth]{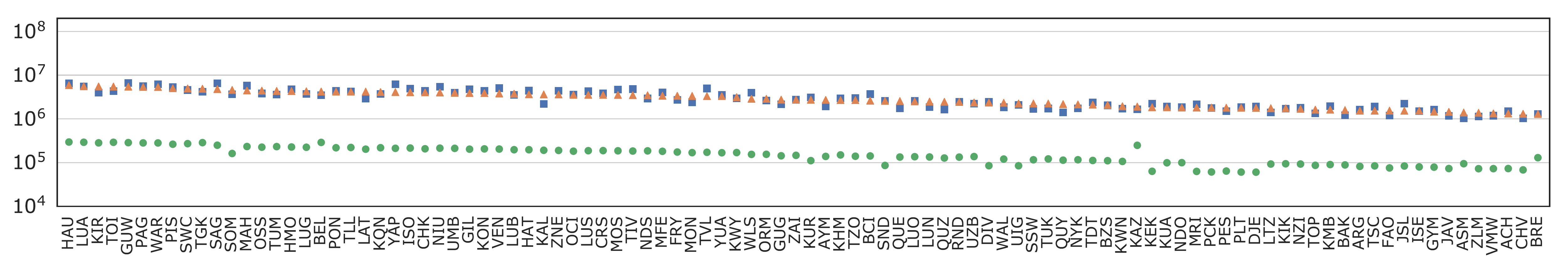}
    \includegraphics[width=\linewidth]{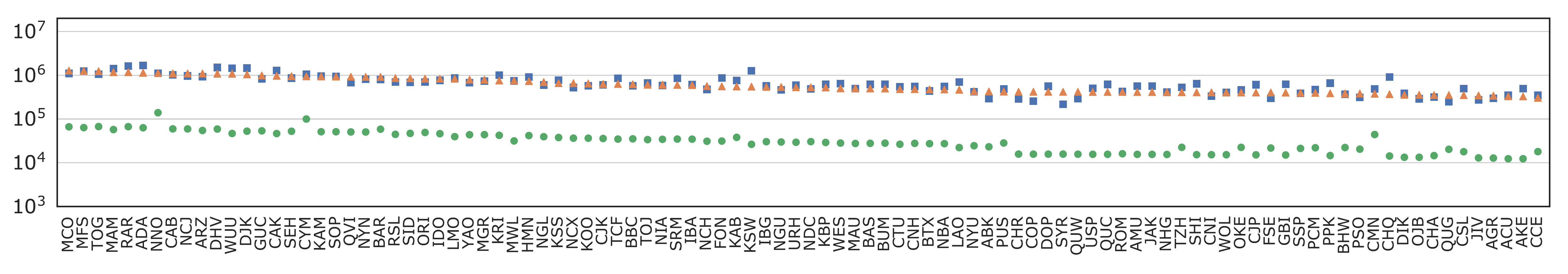}
    \includegraphics[width=\linewidth]{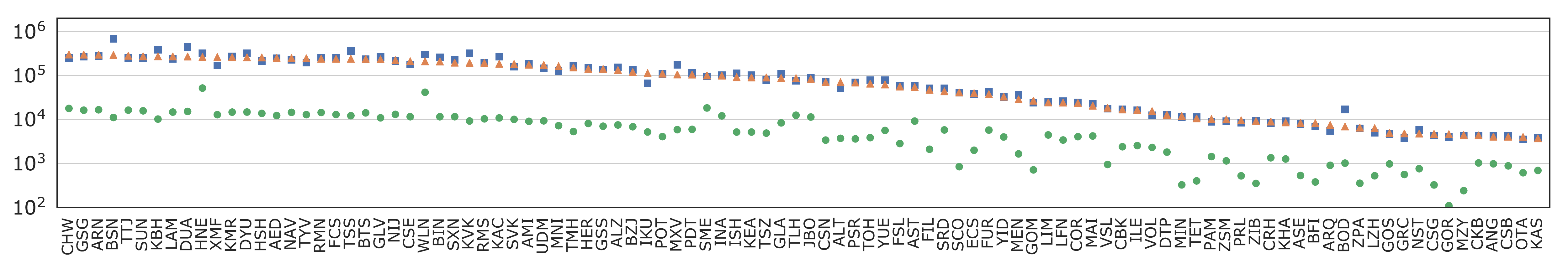}
    \includegraphics[width=\linewidth]{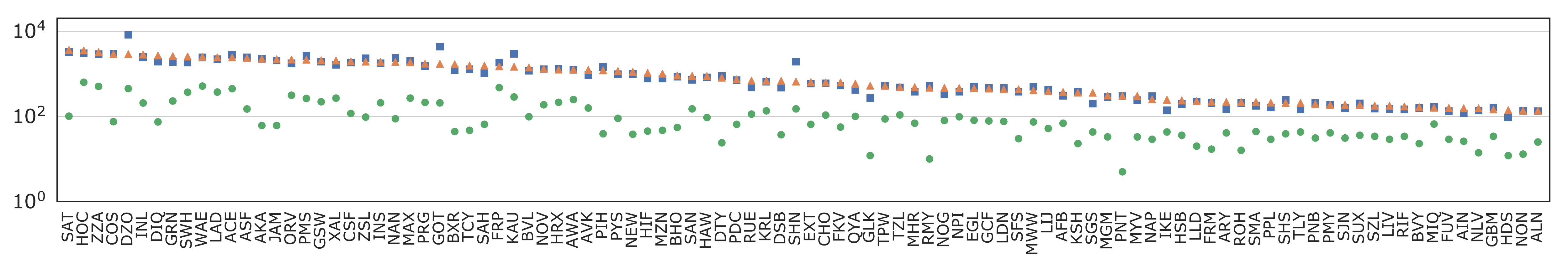}     
    \caption{Training data statistics for the 500 languages, sorted based on descending order of English token count. These statistics are obtained after de-duplication and filtering (see Section~\ref{sec:datasets}). The full name for these ISO 639-3 codes can be looked up using \mtdata, e.g. \texttt{mtdata-iso eng} . }
       \label{fig:train-data-stats}
\end{figure*}

\textbf{Cleaning:}
 We use \textsc{SacreMoses}\footnote{\url{https://github.com/isi-nlp/sacremoses} a fork of \url{https://github.com/alvations/sacremoses} with improvements to tokenization for many low resource languages.} to normalize Unicode punctuations and digits, followed by word tokenization. 
We remove records that are duplicates, have abnormal source-to-target length ratios, have many non-ASCII characters on the English side, have a URL, or which overlap exactly, either on the source or target side, with any sentences in held out sets.
As preprocessing is compute-intensive, we parallelize using Apache Spark.
The cleaning and tokenization results in a corpus of 474 million sentences and 9 billion tokens on the source and English sides each. The token and sentence count for each language are provided in Figure~\ref{fig:train-data-stats}.
Both the processed and raw datasets are available at \url{http://rtg.isi.edu/many-eng/data/v1/}.\footnote{A copy is at \url{https://opus.nlpl.eu/MT560.php}}

\subsection{Many-to-English Multilingual Model}
\label{sec:500eng-model}
\begin{figure*}[ht]
    \centering
    \includegraphics[width=\textwidth]{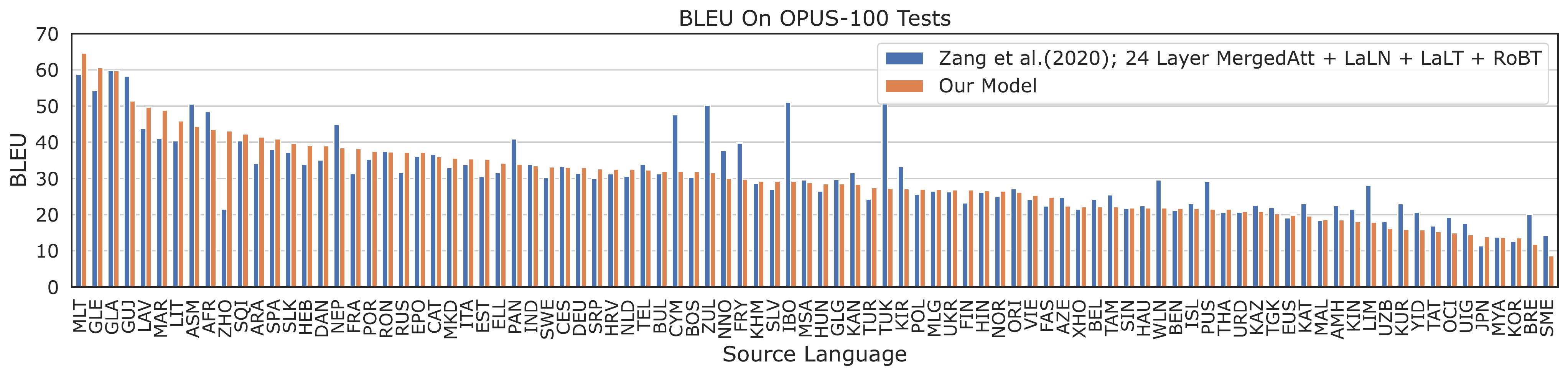}
    \caption[Caption for LOF]{Many-to-English BLEU on OPUS-100 tests~\cite{zhang-etal-2020-multiling-nmt}. 
    Despite having four times more languages on the source side, our model scores competitive BLEU on most languages with the strongest system of \citet{zhang-etal-2020-multiling-nmt}. The tests where our model scores lower BLEU have shorter source sentences (mean length of about three tokens).}
    \label{fig:test-bleu}
\end{figure*}

We use \rtg\ to train Transformer NMT \cite{vaswani2017attention} with a few modifications.
Firstly, instead of a shared BPE vocabulary for both source and target, we use two separate BPE vocabularies. 
Since the source side has 500 languages but the target side has English only, we use a large source vocabulary and a relatively smaller target vocabulary.
A larger target vocabulary leads to higher time and memory complexity, whereas a large source vocabulary increases only the memory complexity but not the time complexity.
We train several models, ranging from the standard 6 layers, 512-dimensional Transformers to larger ones with more parameters. Since the dataset is massive, a larger model trained on big mini-batches yields the best results. Our best performing model is a 768 dimensional model with 12 attention heads, 9 encoder layers, 6 decoder layers, feed-forward dimension of 2048, dropout and label smoothing at 0.1, using $512,000$ and $64,000$ BPE types as source and target vocabularies, respectively. The decoder's input and output embeddings are shared.
Since some of the English sentences are replicated to align with many sentences from different languages (e.g. the Bible corpus), BPE merges are learned from the deduplicated sentences using \nlcodec.
Our best performing model is trained with an effective batch size of about 720,000 tokens per optimizer step. Such big batches are achieved by using mixed-precision distributed training on 8 NVIDIA A100 GPUs with gradient accumulation of 5 mini-batches, each having a maximum of 18,000 tokens. We use the Adam optimizer~\cite{kingma2014adam} with 8000 warm-up steps followed by a decaying learning rate, similar to \citet{vaswani2017attention}. 
We stop training after five days and six hours when a total of 200K updates are made by the optimizer; validation loss is still decreasing at this point. 
To assess the translation quality of our model, we report BLEU~\cite{papineni-etal-2002-bleu,post-2018-sacrebleu}\footnote{All our BLEU scores are obtained from \sacrebleu\ \texttt{BLEU+c.mixed+\#.1+s.exp+tok.13a+v.1.4.13}.} on a subset of languages for which known test sets are available, as given in Figure~\ref{fig:test-bleu}, along with a comparison to \citet{zhang-etal-2020-multiling-nmt}'s best model.\footnote{
Scores are obtained from \url{https://github.com/bzhangGo/zero/tree/master/docs/multilingual_laln_lalt}; accessed: 2021/03/30}

\section{Applications}
\label{sec:value}
The model we trained as a demonstration for our tools is useful on its own, as described in the following sections. 

\subsection{Readily Usable Translation Service}
\label{sec:value.off-shelf-mt}
Our pretrained NMT model is readily usable as a service capable of translating several hundred source languages to English.
By design, source language identification is not necessary.
Figure~\ref{fig:test-bleu} shows that the model scores more than 20 BLEU, which maybe be a useful quality for certain downstream applications involving web and social media content analysis.
Apache Tika \cite{mattmann2011tika}, a content detection and analysis toolkit capable of parsing thousands of file formats, has an option for translating any document into English using our multilingual NMT model.\footnote{\url{https://cwiki.apache.org/confluence/display/TIKA/NMT-RTG}} Our model has been packaged and published to DockerHub\footnote{\url{https://hub.docker.com/}}, which can be obtained by the following command:
\begin{minted}[
%frame=lines,
%framesep=2mm,
baselinestretch=1.1,
fontsize=\footnotesize,
%linenos
]{bash}
IMAGE=tgowda/rtg-model:500toEng-v1
docker run --rm -i -p 6060:6060 $IMAGE 
# For GPU support: --gpus '"device=0"' 
\end{minted}

The above command starts a docker image with HTTP server having a web interface, as can be seen in Figure~\ref{fig:rtg-webui}, and a REST API.
An example interaction with the REST API is as follows: 
\begin{figure}[ht]
    \centering
    \includegraphics[width=\linewidth,trim=20 60 100 15,clip]{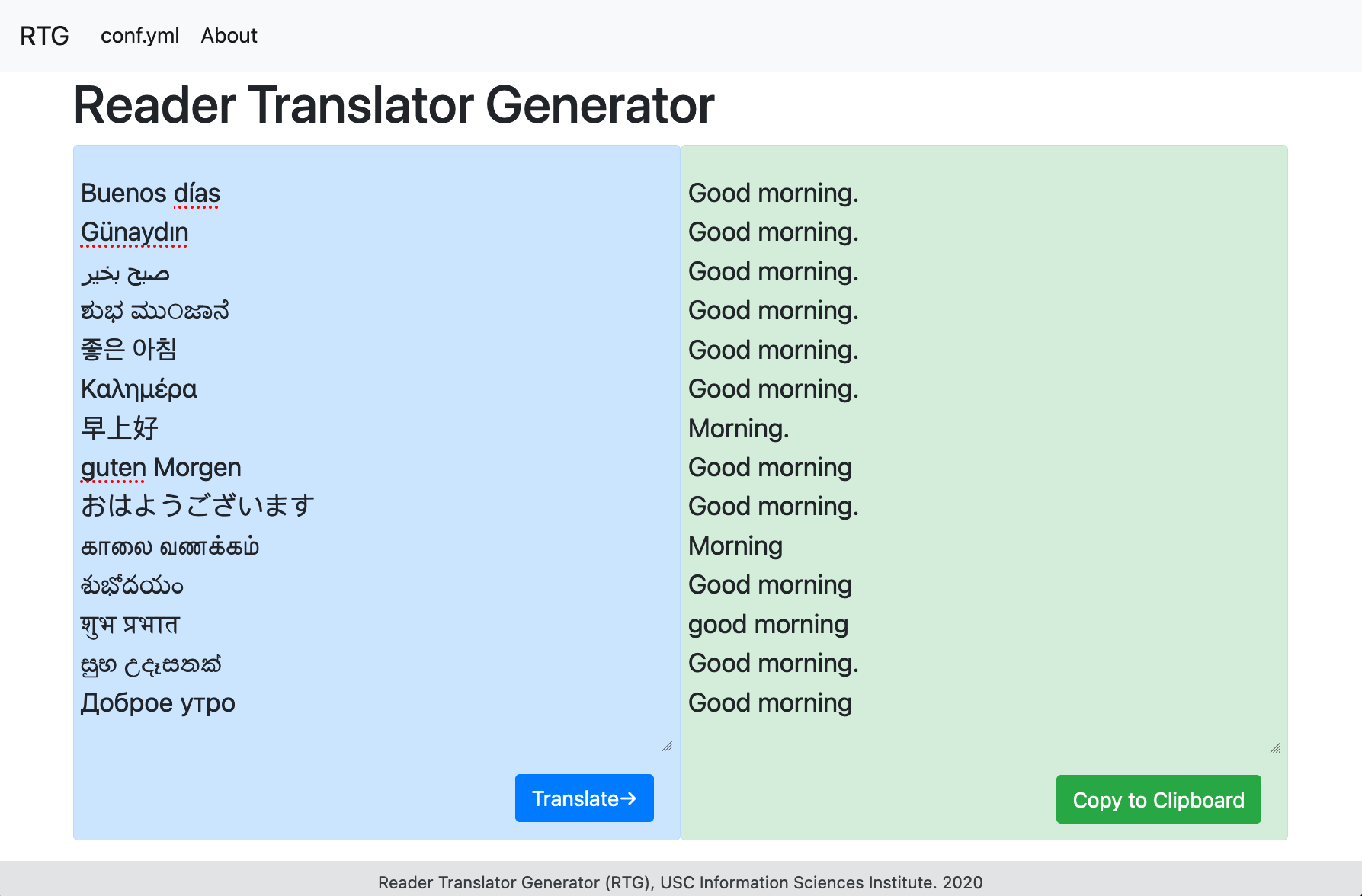}
    \caption{RTG Web Interface}
    \label{fig:rtg-webui}
\end{figure}

\begin{minted}[baselinestretch=1.1, fontsize=\footnotesize]{bash}
curl --data "source=Comment allez-vous?"\
    --data "source=Bonne journée"\
    http://localhost:6060/translate
\end{minted}
\begin{minted}[baselinestretch=1.1, fontsize=\footnotesize]{json}
 {
  "source": [ "Comment allez-vous?",
        "Bonne journée" ],
  "translation": [ "How are you?", 
         "Have a nice day" ]
}
\end{minted}

\subsection{Parent Model for Low Resource MT}
\label{sec:value.transfer-learning}
 Fine tuning is a useful transfer learning technique for improving the translation of low resource languages ~\cite{zoph-etal-2016-transfer,neubig-hu-2018-rapid,gheini2019universal}. 
For instance, consider Breton-English (BRE-ENG) and Northern Sami-English (SME-ENG), two of the low resource settings for which our model has relatively poor BLEU (see Figure~\ref{fig:test-bleu}). 
To show the utility of fine tuning with our model, we train a strong baseline Transformer model, one for each language, from scratch using OPUS-100 training data~\cite{zhang-etal-2020-multiling-nmt}, and finetune our multilingual model on the same dataset as the baselines. We shrink the parent model vocabulary and embeddings to the child model dataset, and train all models on NVIDIA P100 GPUs until convergence.\footnote{More info: \url{https://github.com/thammegowda/006-many-to-eng/tree/master/lowres-xfer}}
 Table~\ref{tab:transfer-lowres}, which shows BLEU on the OPUS-100 test set for the two low resource languages indicates that our multilingual NMT parent model can be further improved with finetuning on limited training data. The finetuned model is significantly better than baseline model.

\begin{table}[ht]
    \centering
    \footnotesize
    \begin{tabular}{l  r r}
        Model     & BRE-ENG & SME-ENG \\ \hline\hline
        Baseline  & 12.7  &  10.7  \\
        Parent    & 11.8  &  8.6  \\
        \textit{Finetuned} & \textbf{22.8}  & \textbf{19.1}  \\ 
    \end{tabular}
    \caption{Finetuning our multilingual NMT on limited training data in low resource settings significantly improves translation quality, as quantified by BLEU.}
    \label{tab:transfer-lowres}
\end{table}
\section{Related work}

\subsection{Tools}
\textsc{SacreBleu}~\cite{post-2018-sacrebleu} simplifies MT evaluation.
\mtdata\ attempts to simplify training setup by automating training and validation dataset retrieval.
\textsc{OPUSTools}~\cite{aulamo-etal-2020-opustools} is a similar tool however, it interfaces with OPUS servers only.
Since the dataset index for \textsc{OPUSTools}~ is on a server, the addition of new datasets requires privileged access.
In contrast, \mtdata\ is a client side library, it can be easily forked and extended to include new datasets without needing special privileges. 

\textbf{\nlcodec:} 
\nlcodec\ is a Python library for vocabulary management. It overcomes the multithreading bottleneck in Python by using PySpark.
\sentpiece~\cite{kudo-richardson-2018-sentencepiece} and \hftok~\cite{wolf-etal-2020-transformers} are the closest alternatives in terms of features, however, modification is relatively difficult for Python users as these libraries are implemented in C++ and Rust, respectively.
In addition, \sentpiece\ uses a binary format for model persistence in favor of efficiency, which takes away the inspectability of the model state. 
Retaining the ability to inspect models and modify core functionality is beneficial for further improving encoding schemes, e.g. subword regularization~\cite{kudo-2018-subwordreg}, BPE dropout~\cite{provilkov-etal-2020-bpedrop}, and optimal stop condition for subword merges~\cite{gowda-may-2020-vocabsize}.
FastBPE is another efficient BPE tool written in C++.\footnote{\url{https://github.com/glample/fastBPE}}  
Subword-nmt~\cite{sennrich-etal-2016-BPE} is a Python implementation of BPE, and stores the model in an inspectable plain text format, however, it is not readily scalable to massive datasets such as the one used in this work.
None of these tools have an equivalent to \nldb's mechanism for efficiently storing and retrieving variable length sequences for distributed training.

\textbf{\rtg:}
Tensor2Tensor \cite{vaswani-etal-2018-tensor2tensor} originally offered the Transformer \cite{vaswani2017attention} implementation using Tensorflow~\cite{tensorflow2015-whitepaper}; 
our implementation uses Pytorch \cite{NEURIPS2019_Pytorch} following \textit{Annotated Transformer} \cite{rush-2018-annotated}.
OpenNMT currently offers separate implementations for both Pytorch and Tensorflow backends~\cite{klein-etal-2017-opennmt,klein-etal-2020-opennmt}.
As open-source toolkits evolve, many good features tend to propagate between them, leading to varying degrees of similarities. Some of the available NMT toolkits are:
Nematus~\cite{sennrich-etal-2017-nematus}, 
xNMT~\cite{neubig-etal-2018-xnmt}.
Marian NMT~\cite{junczys-dowmunt-etal-2018-marian-fast},
Joey NMT~\cite{kreutzer-etal-2019-joeynmt},
Fairseq~\cite{ott-etal-2019-fairseq}, and 
Sockey~\cite{hieber-etal-2020-sockeye}.
An exhaustive comparison of these NMT toolkits is beyond the scope of our current work.

\subsection{Multilingual NMT}
\citet{johnson-etal-2017-googleNMT} show that NMT models are capable of multilingual translation without any architectural changes, and observe that when languages with abundant data are mixed with low resource languages, the translation quality of low resource pairs are significantly improved. They use a private dataset of 12 language pairs; we use publicly available datasets for up to 500 languages. 
\citet{qi-etal-2018-pretrainemb} assemble a multi-parallel dataset for 58 languages from TEDTalks domains, which are included in our dataset. 
\citet{zhang-etal-2020-multiling-nmt} curate OPUS-100, a multilingual dataset of 100 languages sampled from OPUS, including test sets; which are used in this work.
\citet{tiedemann-2020-tatoeba} have established a benchmark task for 500 languages  including single directional baseline models.
\citet{wang-etal-2020-balancing} examine the language-wise imbalance problem in multilingual datasets and propose a method to address the imbalance using a scoring function, which we plan to explore in the future.

\section{Conclusion}

We have introduced our tools: \mtdata\ for downloading datasets, \nlcodec\ for processing, storing and retrieving large scale training data, and \rtg\ for training NMT models.
Using these tools, we have collected a massive dataset and trained a multilingual model for many-to-English translation.
We have demonstrated that our model can be used independently as a translation service, and also showed its use as a parent model for improving low resource language translation. 
All the described tools, used datasets, and trained models are made available to the public for free. 

\section*{Acknowledgments}
The authors would like to thank Lukas Ferrer, Luke Miles, and Mozhdeh Gheini for their contributions to some of the tools used in this work, and thank Jörg Tiedemann for hosting our prepared dataset at OPUS (\url{https://opus.nlpl.eu/MT560.php}). 
The authors acknowledge the Center for Advanced Research Computing (CARC) at the University of Southern California for providing computing resources that have contributed to the research results reported within this publication. URL: \url{https://carc.usc.edu}. 
The authors acknowledge the Texas Advanced Computing Center (TACC) at The University of Texas at Austin for providing HPC resources that have contributed to the research results reported within this paper. URL: \url{http://www.tacc.utexas.edu}. This research is based upon work supported by the Office of the Director of National Intelligence (ODNI), Intelligence Advanced Research Projects Activity (IARPA), via AFRL Contract FA8650-17-C-9116.  The views and conclusions contained herein are those of the authors and should not be interpreted as necessarily representing the official policies or endorsements, either expressed or implied, of the ODNI, IARPA, or the U.S. Government. The U.S. Government is authorized to reproduce and distribute reprints for Governmental purposes notwithstanding any copyright annotation thereon.

\section*{Ethical Consideration}

\textit{Failure Modes:} \mtdata\ will fail to operate, unless patched, when hosting services change their URLs or formats over time.
On certain scenarios when a dataset has been previously accessed and retained in local cache, \mtdata\ continues to operate with a copy of previous version and ignores server side updates.
We have done our best effort in normalizing languages to ISO 639-3 standard; our current version does not accommodate country and script variations of languages; e.g. UK English and US English are both mapped to \textit{eng}. 
Our multilingual NMT model is trained to translate a full sentence at a time without considering source language information; translation of short phrases without a proper context might result in a poor quality translation. 

\textit{Diversity and Fairness:}
We cover all languages on the source side for which publicly available dataset exists, which happens to be about 500 source languages. 
Our model translates to English only, hence only English speakers are benefited from this work.

\textit{Climate Impact:}
\mtdata\ reduces network transfers to the minimal by maintaining a local cache to avoid repetitive downloads.
In addition to the raw datasets, preprocessed data is also available to avoid repetitive computation.
Our Multilingual NMT has higher energy cost than a typical single directional NMT model due to higher number of parameters, however, since our single model translates hundreds of languages, the energy requirement is significantly lower than the total consumption of all independent models. 
Our trained models with all the weights are also made available for download.

\textit{Dataset Ownership:}
\mtdata\ is a client side library that does not have the ownership of datasets in its index.
Addition, removal, or modification in its index is to be submitted by creating an issue at \url{https://github.com/thammegowda/mtdata/issues}. 
We ask the dataset users to review the dataset license, and acknowledge its original creators by citing their work, whose \BibTeX\ entries may be accessed using:\\ \texttt{\footnotesize mtdata list -n <NAME> -l <L1-L2> --full} \\
The prepared dataset that we have made available for download includes \texttt{citations.bib} that acknowledges all the original creators of datasets.
We do not vouch for quality and fairness of all the datasets.

\bibliography{80-refs}

\begin{thebibliography}{57}
\expandafter\ifx\csname natexlab\endcsname\relax\def\natexlab#1{#1}\fi

\bibitem[{Abadi et~al.(2015)Abadi, Agarwal, Barham, Brevdo, Chen, Citro,
  Corrado, Davis, Dean, Devin, Ghemawat, Goodfellow, Harp, Irving, Isard, Jia,
  Jozefowicz, Kaiser, Kudlur, Levenberg, Man\'{e}, Monga, Moore, Murray, Olah,
  Schuster, Shlens, Steiner, Sutskever, Talwar, Tucker, Vanhoucke, Vasudevan,
  Vi\'{e}gas, Vinyals, Warden, Wattenberg, Wicke, Yu, and
  Zheng}]{tensorflow2015-whitepaper}
Mart\'{\i}n Abadi, Ashish Agarwal, Paul Barham, Eugene Brevdo, Zhifeng Chen,
  Craig Citro, Greg~S. Corrado, Andy Davis, Jeffrey Dean, Matthieu Devin,
  Sanjay Ghemawat, Ian Goodfellow, Andrew Harp, Geoffrey Irving, Michael Isard,
  Yangqing Jia, Rafal Jozefowicz, Lukasz Kaiser, Manjunath Kudlur, Josh
  Levenberg, Dandelion Man\'{e}, Rajat Monga, Sherry Moore, Derek Murray, Chris
  Olah, Mike Schuster, Jonathon Shlens, Benoit Steiner, Ilya Sutskever, Kunal
  Talwar, Paul Tucker, Vincent Vanhoucke, Vijay Vasudevan, Fernanda Vi\'{e}gas,
  Oriol Vinyals, Pete Warden, Martin Wattenberg, Martin Wicke, Yuan Yu, and
  Xiaoqiang Zheng. 2015.
\newblock \href {https://www.tensorflow.org/} {{TensorFlow}: Large-scale
  machine learning on heterogeneous systems}.
\newblock Software available from tensorflow.org.

\bibitem[{Agi{\'c} and Vuli{\'c}(2019)}]{agic-vulic-2019-jw300}
{\v{Z}}eljko Agi{\'c} and Ivan Vuli{\'c}. 2019.
\newblock \href {https://doi.org/10.18653/v1/P19-1310} {{JW}300: A
  wide-coverage parallel corpus for low-resource languages}.
\newblock In \emph{Proceedings of the 57th Annual Meeting of the Association
  for Computational Linguistics}, pages 3204--3210, Florence, Italy.
  Association for Computational Linguistics.

\bibitem[{Aulamo et~al.(2020)Aulamo, Sulubacak, Virpioja, and
  Tiedemann}]{aulamo-etal-2020-opustools}
Mikko Aulamo, Umut Sulubacak, Sami Virpioja, and J{\"o}rg Tiedemann. 2020.
\newblock \href {https://www.aclweb.org/anthology/2020.lrec-1.467}
  {{O}pus{T}ools and parallel corpus diagnostics}.
\newblock In \emph{Proceedings of the 12th Language Resources and Evaluation
  Conference}, pages 3782--3789, Marseille, France. European Language Resources
  Association.

\bibitem[{Bahdanau et~al.(2015)Bahdanau, Cho, and Bengio}]{bahdanau2014nmtattn}
Dzmitry Bahdanau, Kyunghyun Cho, and Yoshua Bengio. 2015.
\newblock \href {http://arxiv.org/abs/1409.0473} {Neural machine translation by
  jointly learning to align and translate}.
\newblock In \emph{3rd International Conference on Learning Representations,
  {ICLR} 2015, San Diego, CA, USA, May 7-9, 2015, Conference Track
  Proceedings}.

\bibitem[{Barrault et~al.(2020)Barrault, Biesialska, Bojar, Costa-juss{\`a},
  Federmann, Graham, Grundkiewicz, Haddow, Huck, Joanis, Kocmi, Koehn, Lo,
  Ljube{\v{s}}i{\'c}, Monz, Morishita, Nagata, Nakazawa, Pal, Post, and
  Zampieri}]{barrault-etal-2020-findings}
Lo{\"\i}c Barrault, Magdalena Biesialska, Ond{\v{r}}ej Bojar, Marta~R.
  Costa-juss{\`a}, Christian Federmann, Yvette Graham, Roman Grundkiewicz,
  Barry Haddow, Matthias Huck, Eric Joanis, Tom Kocmi, Philipp Koehn, Chi-kiu
  Lo, Nikola Ljube{\v{s}}i{\'c}, Christof Monz, Makoto Morishita, Masaaki
  Nagata, Toshiaki Nakazawa, Santanu Pal, Matt Post, and Marcos Zampieri. 2020.
\newblock \href {https://www.aclweb.org/anthology/2020.wmt-1.1} {Findings of
  the 2020 conference on machine translation ({WMT}20)}.
\newblock In \emph{Proceedings of the Fifth Conference on Machine Translation},
  pages 1--55, Online. Association for Computational Linguistics.

\bibitem[{Barrault et~al.(2019)Barrault, Bojar, Costa-juss{\`a}, Federmann,
  Fishel, Graham, Haddow, Huck, Koehn, Malmasi, Monz, M{\"u}ller, Pal, Post,
  and Zampieri}]{barrault-etal-2019-findings}
Lo{\"\i}c Barrault, Ond{\v{r}}ej Bojar, Marta~R. Costa-juss{\`a}, Christian
  Federmann, Mark Fishel, Yvette Graham, Barry Haddow, Matthias Huck, Philipp
  Koehn, Shervin Malmasi, Christof Monz, Mathias M{\"u}ller, Santanu Pal, Matt
  Post, and Marcos Zampieri. 2019.
\newblock \href {https://doi.org/10.18653/v1/W19-5301} {Findings of the 2019
  conference on machine translation ({WMT}19)}.
\newblock In \emph{Proceedings of the Fourth Conference on Machine Translation
  (Volume 2: Shared Task Papers, Day 1)}, pages 1--61, Florence, Italy.
  Association for Computational Linguistics.

\bibitem[{Bojar et~al.(2013)Bojar, Buck, Callison-Burch, Federmann, Haddow,
  Koehn, Monz, Post, Soricut, and Specia}]{bojar-etal-2013-findings}
Ond{\v{r}}ej Bojar, Christian Buck, Chris Callison-Burch, Christian Federmann,
  Barry Haddow, Philipp Koehn, Christof Monz, Matt Post, Radu Soricut, and
  Lucia Specia. 2013.
\newblock \href {https://www.aclweb.org/anthology/W13-2201} {Findings of the
  2013 {W}orkshop on {S}tatistical {M}achine {T}ranslation}.
\newblock In \emph{Proceedings of the Eighth Workshop on Statistical Machine
  Translation}, pages 1--44, Sofia, Bulgaria. Association for Computational
  Linguistics.

\bibitem[{Bojar et~al.(2014)Bojar, Buck, Federmann, Haddow, Koehn, Leveling,
  Monz, Pecina, Post, Saint-Amand, Soricut, Specia, and
  Tamchyna}]{bojar-etal-2014-findings}
Ond{\v{r}}ej Bojar, Christian Buck, Christian Federmann, Barry Haddow, Philipp
  Koehn, Johannes Leveling, Christof Monz, Pavel Pecina, Matt Post, Herve
  Saint-Amand, Radu Soricut, Lucia Specia, and Ale{\v{s}} Tamchyna. 2014.
\newblock \href {https://doi.org/10.3115/v1/W14-3302} {Findings of the 2014
  workshop on statistical machine translation}.
\newblock In \emph{Proceedings of the Ninth Workshop on Statistical Machine
  Translation}, pages 12--58, Baltimore, Maryland, USA. Association for
  Computational Linguistics.

\bibitem[{Bojar et~al.(2017)Bojar, Chatterjee, Federmann, Graham, Haddow,
  Huang, Huck, Koehn, Liu, Logacheva, Monz, Negri, Post, Rubino, Specia, and
  Turchi}]{bojar-etal-2017-findings}
Ond{\v{r}}ej Bojar, Rajen Chatterjee, Christian Federmann, Yvette Graham, Barry
  Haddow, Shujian Huang, Matthias Huck, Philipp Koehn, Qun Liu, Varvara
  Logacheva, Christof Monz, Matteo Negri, Matt Post, Raphael Rubino, Lucia
  Specia, and Marco Turchi. 2017.
\newblock \href {https://doi.org/10.18653/v1/W17-4717} {Findings of the 2017
  conference on machine translation ({WMT}17)}.
\newblock In \emph{Proceedings of the Second Conference on Machine
  Translation}, pages 169--214, Copenhagen, Denmark. Association for
  Computational Linguistics.

\bibitem[{Bojar et~al.(2016)Bojar, Chatterjee, Federmann, Graham, Haddow, Huck,
  Jimeno~Yepes, Koehn, Logacheva, Monz, Negri, N{\'e}v{\'e}ol, Neves, Popel,
  Post, Rubino, Scarton, Specia, Turchi, Verspoor, and
  Zampieri}]{bojar-etal-2016-findings}
Ond{\v{r}}ej Bojar, Rajen Chatterjee, Christian Federmann, Yvette Graham, Barry
  Haddow, Matthias Huck, Antonio Jimeno~Yepes, Philipp Koehn, Varvara
  Logacheva, Christof Monz, Matteo Negri, Aur{\'e}lie N{\'e}v{\'e}ol, Mariana
  Neves, Martin Popel, Matt Post, Raphael Rubino, Carolina Scarton, Lucia
  Specia, Marco Turchi, Karin Verspoor, and Marcos Zampieri. 2016.
\newblock \href {https://doi.org/10.18653/v1/W16-2301} {Findings of the 2016
  conference on machine translation}.
\newblock In \emph{Proceedings of the First Conference on Machine Translation:
  Volume 2, Shared Task Papers}, pages 131--198, Berlin, Germany. Association
  for Computational Linguistics.

\bibitem[{Bojar et~al.(2015)Bojar, Chatterjee, Federmann, Haddow, Huck, Hokamp,
  Koehn, Logacheva, Monz, Negri, Post, Scarton, Specia, and
  Turchi}]{bojar-etal-2015-findings}
Ond{\v{r}}ej Bojar, Rajen Chatterjee, Christian Federmann, Barry Haddow,
  Matthias Huck, Chris Hokamp, Philipp Koehn, Varvara Logacheva, Christof Monz,
  Matteo Negri, Matt Post, Carolina Scarton, Lucia Specia, and Marco Turchi.
  2015.
\newblock \href {https://doi.org/10.18653/v1/W15-3001} {Findings of the 2015
  workshop on statistical machine translation}.
\newblock In \emph{Proceedings of the Tenth Workshop on Statistical Machine
  Translation}, pages 1--46, Lisbon, Portugal. Association for Computational
  Linguistics.

\bibitem[{Bojar et~al.(2018)Bojar, Federmann, Fishel, Graham, Haddow, Koehn,
  and Monz}]{bojar-etal-2018-findings}
Ond{\v{r}}ej Bojar, Christian Federmann, Mark Fishel, Yvette Graham, Barry
  Haddow, Philipp Koehn, and Christof Monz. 2018.
\newblock \href {https://doi.org/10.18653/v1/W18-6401} {Findings of the 2018
  conference on machine translation ({WMT}18)}.
\newblock In \emph{Proceedings of the Third Conference on Machine Translation:
  Shared Task Papers}, pages 272--303, Belgium, Brussels. Association for
  Computational Linguistics.

\bibitem[{Espl{\`a} et~al.(2019)Espl{\`a}, Forcada, Ram{\'\i}rez-S{\'a}nchez,
  and Hoang}]{espla-etal-2019-paracrawl}
Miquel Espl{\`a}, Mikel Forcada, Gema Ram{\'\i}rez-S{\'a}nchez, and Hieu Hoang.
  2019.
\newblock \href {https://www.aclweb.org/anthology/W19-6721} {{P}ara{C}rawl:
  Web-scale parallel corpora for the languages of the {EU}}.
\newblock In \emph{Proceedings of Machine Translation Summit XVII Volume 2:
  Translator, Project and User Tracks}, pages 118--119, Dublin, Ireland.
  European Association for Machine Translation.

\bibitem[{Gheini and May(2019)}]{gheini2019universal}
Mozhdeh Gheini and Jonathan May. 2019.
\newblock A universal parent model for low-resource neural machine translation
  transfer.
\newblock \emph{arXiv preprint arXiv:1909.06516}.

\bibitem[{Gowda and May(2020)}]{gowda-may-2020-vocabsize}
Thamme Gowda and Jonathan May. 2020.
\newblock \href {https://doi.org/10.18653/v1/2020.findings-emnlp.352} {Finding
  the optimal vocabulary size for neural machine translation}.
\newblock In \emph{Findings of the Association for Computational Linguistics:
  EMNLP 2020}, pages 3955--3964, Online. Association for Computational
  Linguistics.

\bibitem[{Haddow and Kirefu(2020)}]{haddow2020pmindia}
Barry Haddow and Faheem Kirefu. 2020.
\newblock \href {http://arxiv.org/abs/2001.09907} {Pmindia -- a collection of
  parallel corpora of languages of india}.

\bibitem[{Harris et~al.(2020)Harris, Millman, van~der Walt, Gommers, Virtanen,
  Cournapeau, Wieser, Taylor, Berg, Smith, Kern, Picus, Hoyer, van Kerkwijk,
  Brett, Haldane, del R{'{\i}}o, Wiebe, Peterson, G{'{e}}rard-Marchant,
  Sheppard, Reddy, Weckesser, Abbasi, Gohlke, and Oliphant}]{harris2020numpy}
Charles~R. Harris, K.~Jarrod Millman, St{'{e}}fan~J. van~der Walt, Ralf
  Gommers, Pauli Virtanen, David Cournapeau, Eric Wieser, Julian Taylor,
  Sebastian Berg, Nathaniel~J. Smith, Robert Kern, Matti Picus, Stephan Hoyer,
  Marten~H. van Kerkwijk, Matthew Brett, Allan Haldane, Jaime~Fern{'{a}}ndez
  del R{'{\i}}o, Mark Wiebe, Pearu Peterson, Pierre G{'{e}}rard-Marchant, Kevin
  Sheppard, Tyler Reddy, Warren Weckesser, Hameer Abbasi, Christoph Gohlke, and
  Travis~E. Oliphant. 2020.
\newblock \href {https://doi.org/10.1038/s41586-020-2649-2} {Array programming
  with {NumPy}}.
\newblock \emph{Nature}, 585(7825):357--362.

\bibitem[{Hieber et~al.(2020)Hieber, Domhan, Denkowski, and
  Vilar}]{hieber-etal-2020-sockeye}
Felix Hieber, Tobias Domhan, Michael Denkowski, and David Vilar. 2020.
\newblock \href {https://www.aclweb.org/anthology/2020.eamt-1.50} {Sockeye 2: A
  toolkit for neural machine translation}.
\newblock In \emph{Proceedings of the 22nd Annual Conference of the European
  Association for Machine Translation}, pages 457--458, Lisboa, Portugal.
  European Association for Machine Translation.

\bibitem[{Johnson et~al.(2017)Johnson, Schuster, Le, Krikun, Wu, Chen, Thorat,
  Vi{\'e}gas, Wattenberg, Corrado, Hughes, and
  Dean}]{johnson-etal-2017-googleNMT}
Melvin Johnson, Mike Schuster, Quoc~V. Le, Maxim Krikun, Yonghui Wu, Zhifeng
  Chen, Nikhil Thorat, Fernanda Vi{\'e}gas, Martin Wattenberg, Greg Corrado,
  Macduff Hughes, and Jeffrey Dean. 2017.
\newblock \href {https://doi.org/10.1162/tacl_a_00065} {{G}oogle{'}s
  multilingual neural machine translation system: Enabling zero-shot
  translation}.
\newblock \emph{Transactions of the Association for Computational Linguistics},
  5:339--351.

\bibitem[{Junczys-Dowmunt et~al.(2018)Junczys-Dowmunt, Grundkiewicz, Dwojak,
  Hoang, Heafield, Neckermann, Seide, Germann, Aji, Bogoychev, Martins, and
  Birch}]{junczys-dowmunt-etal-2018-marian-fast}
Marcin Junczys-Dowmunt, Roman Grundkiewicz, Tomasz Dwojak, Hieu Hoang, Kenneth
  Heafield, Tom Neckermann, Frank Seide, Ulrich Germann, Alham~Fikri Aji,
  Nikolay Bogoychev, Andr{\'e} F.~T. Martins, and Alexandra Birch. 2018.
\newblock \href {https://doi.org/10.18653/v1/P18-4020} {{M}arian: Fast neural
  machine translation in {C}++}.
\newblock In \emph{Proceedings of {ACL} 2018, System Demonstrations}, pages
  116--121, Melbourne, Australia. Association for Computational Linguistics.

\bibitem[{Kingma and Ba(2014)}]{kingma2014adam}
Diederik~P Kingma and Jimmy Ba. 2014.
\newblock Adam: A method for stochastic optimization.
\newblock \emph{arXiv preprint arXiv:1412.6980}.

\bibitem[{Klein et~al.(2020)Klein, Hernandez, Nguyen, and
  Senellart}]{klein-etal-2020-opennmt}
Guillaume Klein, Fran{\c{c}}ois Hernandez, Vincent Nguyen, and Jean Senellart.
  2020.
\newblock \href {https://www.aclweb.org/anthology/2020.amta-research.9} {The
  {O}pen{NMT} neural machine translation toolkit: 2020 edition}.
\newblock In \emph{Proceedings of the 14th Conference of the Association for
  Machine Translation in the Americas (Volume 1: Research Track)}, pages
  102--109, Virtual. Association for Machine Translation in the Americas.

\bibitem[{Klein et~al.(2017)Klein, Kim, Deng, Senellart, and
  Rush}]{klein-etal-2017-opennmt}
Guillaume Klein, Yoon Kim, Yuntian Deng, Jean Senellart, and Alexander Rush.
  2017.
\newblock \href {https://www.aclweb.org/anthology/P17-4012} {{O}pen{NMT}:
  Open-source toolkit for neural machine translation}.
\newblock In \emph{Proceedings of {ACL} 2017, System Demonstrations}, pages
  67--72, Vancouver, Canada. Association for Computational Linguistics.

\bibitem[{Koehn(2005)}]{koehn2005europarl}
Philipp Koehn. 2005.
\newblock Europarl: A parallel corpus for statistical machine translation.
\newblock \emph{Proc. 10th Machine Translation Summit (MT Summit), 2005}, pages
  79--86.

\bibitem[{Kreutzer et~al.(2019)Kreutzer, Bastings, and
  Riezler}]{kreutzer-etal-2019-joeynmt}
Julia Kreutzer, Jasmijn Bastings, and Stefan Riezler. 2019.
\newblock \href {https://doi.org/10.18653/v1/D19-3019} {Joey {NMT}: A
  minimalist {NMT} toolkit for novices}.
\newblock In \emph{Proceedings of the 2019 Conference on Empirical Methods in
  Natural Language Processing and the 9th International Joint Conference on
  Natural Language Processing (EMNLP-IJCNLP): System Demonstrations}, pages
  109--114, Hong Kong, China. Association for Computational Linguistics.

\bibitem[{Kudo(2018)}]{kudo-2018-subwordreg}
Taku Kudo. 2018.
\newblock \href {https://doi.org/10.18653/v1/P18-1007} {Subword regularization:
  Improving neural network translation models with multiple subword
  candidates}.
\newblock In \emph{Proceedings of the 56th Annual Meeting of the Association
  for Computational Linguistics (Volume 1: Long Papers)}, pages 66--75,
  Melbourne, Australia. Association for Computational Linguistics.

\bibitem[{Kudo and Richardson(2018)}]{kudo-richardson-2018-sentencepiece}
Taku Kudo and John Richardson. 2018.
\newblock \href {https://doi.org/10.18653/v1/D18-2012} {{S}entence{P}iece: A
  simple and language independent subword tokenizer and detokenizer for neural
  text processing}.
\newblock In \emph{Proceedings of the 2018 Conference on Empirical Methods in
  Natural Language Processing: System Demonstrations}, pages 66--71, Brussels,
  Belgium. Association for Computational Linguistics.

\bibitem[{Kunchukuttan et~al.(2018)Kunchukuttan, Mehta, and
  Bhattacharyya}]{kunchukuttan-etal-2018-iit}
Anoop Kunchukuttan, Pratik Mehta, and Pushpak Bhattacharyya. 2018.
\newblock \href {https://www.aclweb.org/anthology/L18-1548} {The {IIT} {B}ombay
  {E}nglish-{H}indi parallel corpus}.
\newblock In \emph{Proceedings of the Eleventh International Conference on
  Language Resources and Evaluation ({LREC} 2018)}, Miyazaki, Japan. European
  Language Resources Association (ELRA).

\bibitem[{Luong et~al.(2015)Luong, Pham, and Manning}]{luong2015effectiveAttn}
Thang Luong, Hieu Pham, and Christopher~D. Manning. 2015.
\newblock \href {https://doi.org/10.18653/v1/D15-1166} {Effective approaches to
  attention-based neural machine translation}.
\newblock In \emph{Proceedings of the 2015 Conference on Empirical Methods in
  Natural Language Processing}, pages 1412--1421, Lisbon, Portugal. Association
  for Computational Linguistics.

\bibitem[{Mattmann and Zitting(2011)}]{mattmann2011tika}
Chris Mattmann and Jukka Zitting. 2011.
\newblock Tika in action.

\bibitem[{Neubig(2011)}]{neubig11kftt}
Graham Neubig. 2011.
\newblock The {Kyoto} free translation task.
\newblock http://www.phontron.com/kftt.

\bibitem[{Neubig and Hu(2018)}]{neubig-hu-2018-rapid}
Graham Neubig and Junjie Hu. 2018.
\newblock \href {https://doi.org/10.18653/v1/D18-1103} {Rapid adaptation of
  neural machine translation to new languages}.
\newblock In \emph{Proceedings of the 2018 Conference on Empirical Methods in
  Natural Language Processing}, pages 875--880, Brussels, Belgium. Association
  for Computational Linguistics.

\bibitem[{Neubig et~al.(2018)Neubig, Sperber, Wang, Felix, Matthews,
  Padmanabhan, Qi, Sachan, Arthur, Godard, Hewitt, Riad, and
  Wang}]{neubig-etal-2018-xnmt}
Graham Neubig, Matthias Sperber, Xinyi Wang, Matthieu Felix, Austin Matthews,
  Sarguna Padmanabhan, Ye~Qi, Devendra Sachan, Philip Arthur, Pierre Godard,
  John Hewitt, Rachid Riad, and Liming Wang. 2018.
\newblock \href {https://www.aclweb.org/anthology/W18-1818} {{XNMT}: The
  e{X}tensible neural machine translation toolkit}.
\newblock In \emph{Proceedings of the 13th Conference of the Association for
  Machine Translation in the {A}mericas (Volume 1: Research Track)}, pages
  185--192, Boston, MA. Association for Machine Translation in the Americas.

\bibitem[{Ott et~al.(2019)Ott, Edunov, Baevski, Fan, Gross, Ng, Grangier, and
  Auli}]{ott-etal-2019-fairseq}
Myle Ott, Sergey Edunov, Alexei Baevski, Angela Fan, Sam Gross, Nathan Ng,
  David Grangier, and Michael Auli. 2019.
\newblock \href {https://doi.org/10.18653/v1/N19-4009} {fairseq: A fast,
  extensible toolkit for sequence modeling}.
\newblock In \emph{Proceedings of the 2019 Conference of the North {A}merican
  Chapter of the Association for Computational Linguistics (Demonstrations)},
  pages 48--53, Minneapolis, Minnesota. Association for Computational
  Linguistics.

\bibitem[{Papineni et~al.(2002)Papineni, Roukos, Ward, and
  Zhu}]{papineni-etal-2002-bleu}
Kishore Papineni, Salim Roukos, Todd Ward, and Wei-Jing Zhu. 2002.
\newblock \href {https://doi.org/10.3115/1073083.1073135} {{B}leu: a method for
  automatic evaluation of machine translation}.
\newblock In \emph{Proceedings of the 40th Annual Meeting of the Association
  for Computational Linguistics}, pages 311--318, Philadelphia, Pennsylvania,
  USA. Association for Computational Linguistics.

\bibitem[{Paszke et~al.(2019)Paszke, Gross, Massa, Lerer, Bradbury, Chanan,
  Killeen, Lin, Gimelshein, Antiga, Desmaison, Kopf, Yang, DeVito, Raison,
  Tejani, Chilamkurthy, Steiner, Fang, Bai, and Chintala}]{NEURIPS2019_Pytorch}
Adam Paszke, Sam Gross, Francisco Massa, Adam Lerer, James Bradbury, Gregory
  Chanan, Trevor Killeen, Zeming Lin, Natalia Gimelshein, Luca Antiga, Alban
  Desmaison, Andreas Kopf, Edward Yang, Zachary DeVito, Martin Raison, Alykhan
  Tejani, Sasank Chilamkurthy, Benoit Steiner, Lu~Fang, Junjie Bai, and Soumith
  Chintala. 2019.
\newblock \href
  {http://papers.neurips.cc/paper/9015-pytorch-an-imperative-style-high-performance-deep-learning-library.pdf}
  {Pytorch: An imperative style, high-performance deep learning library}.
\newblock In H.~Wallach, H.~Larochelle, A.~Beygelzimer, F.~d\textquotesingle
  Alch\'{e}-Buc, E.~Fox, and R.~Garnett, editors, \emph{Advances in Neural
  Information Processing Systems 32}, pages 8024--8035. Curran Associates, Inc.

\bibitem[{Post(2018)}]{post-2018-sacrebleu}
Matt Post. 2018.
\newblock \href {https://doi.org/10.18653/v1/W18-6319} {A call for clarity in
  reporting {BLEU} scores}.
\newblock In \emph{Proceedings of the Third Conference on Machine Translation:
  Research Papers}, pages 186--191, Brussels, Belgium. Association for
  Computational Linguistics.

\bibitem[{Post et~al.(2012)Post, Callison-Burch, and
  Osborne}]{post-etal-2012-constructing}
Matt Post, Chris Callison-Burch, and Miles Osborne. 2012.
\newblock \href {https://www.aclweb.org/anthology/W12-3152} {Constructing
  parallel corpora for six {I}ndian languages via crowdsourcing}.
\newblock In \emph{Proceedings of the Seventh Workshop on Statistical Machine
  Translation}, pages 401--409, Montr{\'e}al, Canada. Association for
  Computational Linguistics.

\bibitem[{Press and Wolf(2017)}]{press-wolf-2017-embeddings}
Ofir Press and Lior Wolf. 2017.
\newblock \href {https://www.aclweb.org/anthology/E17-2025} {Using the output
  embedding to improve language models}.
\newblock In \emph{Proceedings of the 15th Conference of the {E}uropean Chapter
  of the Association for Computational Linguistics: Volume 2, Short Papers},
  pages 157--163, Valencia, Spain. Association for Computational Linguistics.

\bibitem[{Provilkov et~al.(2020)Provilkov, Emelianenko, and
  Voita}]{provilkov-etal-2020-bpedrop}
Ivan Provilkov, Dmitrii Emelianenko, and Elena Voita. 2020.
\newblock \href {https://doi.org/10.18653/v1/2020.acl-main.170} {{BPE}-dropout:
  Simple and effective subword regularization}.
\newblock In \emph{Proceedings of the 58th Annual Meeting of the Association
  for Computational Linguistics}, pages 1882--1892, Online. Association for
  Computational Linguistics.

\bibitem[{Qi et~al.(2018)Qi, Sachan, Felix, Padmanabhan, and
  Neubig}]{qi-etal-2018-pretrainemb}
Ye~Qi, Devendra Sachan, Matthieu Felix, Sarguna Padmanabhan, and Graham Neubig.
  2018.
\newblock \href {https://doi.org/10.18653/v1/N18-2084} {When and why are
  pre-trained word embeddings useful for neural machine translation?}
\newblock In \emph{Proceedings of the 2018 Conference of the North {A}merican
  Chapter of the Association for Computational Linguistics: Human Language
  Technologies, Volume 2 (Short Papers)}, pages 529--535, New Orleans,
  Louisiana. Association for Computational Linguistics.

\bibitem[{Rozis and Skadi{\c{n}}{\v{s}}(2017)}]{rozis-skadins-2017-tilde}
Roberts Rozis and Raivis Skadi{\c{n}}{\v{s}}. 2017.
\newblock \href {https://www.aclweb.org/anthology/W17-0235} {Tilde {MODEL} -
  multilingual open data for {EU} languages}.
\newblock In \emph{Proceedings of the 21st Nordic Conference on Computational
  Linguistics}, pages 263--265, Gothenburg, Sweden. Association for
  Computational Linguistics.

\bibitem[{Rush(2018)}]{rush-2018-annotated}
Alexander Rush. 2018.
\newblock \href {https://doi.org/10.18653/v1/W18-2509} {The annotated
  transformer}.
\newblock In \emph{Proceedings of Workshop for {NLP} Open Source Software
  ({NLP}-{OSS})}, pages 52--60, Melbourne, Australia. Association for
  Computational Linguistics.

\bibitem[{Schwenk et~al.(2019)Schwenk, Chaudhary, Sun, Gong, and
  Guzm{\'{a}}n}]{schwenk-etal-2019-wikimatrixv1}
Holger Schwenk, Vishrav Chaudhary, Shuo Sun, Hongyu Gong, and Francisco
  Guzm{\'{a}}n. 2019.
\newblock \href {http://arxiv.org/abs/1907.05791} {Wikimatrix: Mining 135m
  parallel sentences in 1620 language pairs from wikipedia}.
\newblock \emph{CoRR}, abs/1907.05791.

\bibitem[{Sennrich et~al.(2017)Sennrich, Firat, Cho, Birch, Haddow, Hitschler,
  Junczys-Dowmunt, L{\"a}ubli, Miceli~Barone, Mokry, and
  N{\u{a}}dejde}]{sennrich-etal-2017-nematus}
Rico Sennrich, Orhan Firat, Kyunghyun Cho, Alexandra Birch, Barry Haddow,
  Julian Hitschler, Marcin Junczys-Dowmunt, Samuel L{\"a}ubli, Antonio~Valerio
  Miceli~Barone, Jozef Mokry, and Maria N{\u{a}}dejde. 2017.
\newblock \href {https://www.aclweb.org/anthology/E17-3017} {{N}ematus: a
  toolkit for neural machine translation}.
\newblock In \emph{Proceedings of the Software Demonstrations of the 15th
  Conference of the {E}uropean Chapter of the Association for Computational
  Linguistics}, pages 65--68, Valencia, Spain. Association for Computational
  Linguistics.

\bibitem[{Sennrich et~al.(2016)Sennrich, Haddow, and
  Birch}]{sennrich-etal-2016-BPE}
Rico Sennrich, Barry Haddow, and Alexandra Birch. 2016.
\newblock \href {https://doi.org/10.18653/v1/P16-1162} {Neural machine
  translation of rare words with subword units}.
\newblock In \emph{Proceedings of the 54th Annual Meeting of the Association
  for Computational Linguistics (Volume 1: Long Papers)}, pages 1715--1725,
  Berlin, Germany. Association for Computational Linguistics.

\bibitem[{Tiedemann(2012)}]{tiedemann-2012-parallel}
J{\"o}rg Tiedemann. 2012.
\newblock \href
  {http://www.lrec-conf.org/proceedings/lrec2012/pdf/463_Paper.pdf} {Parallel
  data, tools and interfaces in {OPUS}}.
\newblock In \emph{Proceedings of the Eighth International Conference on
  Language Resources and Evaluation ({LREC}'12)}, pages 2214--2218, Istanbul,
  Turkey. European Language Resources Association (ELRA).

\bibitem[{Tiedemann(2020)}]{tiedemann-2020-tatoeba}
J{\"o}rg Tiedemann. 2020.
\newblock \href {https://www.aclweb.org/anthology/2020.wmt-1.139} {The tatoeba
  translation challenge {--} realistic data sets for low resource and
  multilingual {MT}}.
\newblock In \emph{Proceedings of the Fifth Conference on Machine Translation},
  pages 1174--1182, Online. Association for Computational Linguistics.

\bibitem[{Vaswani et~al.(2018)Vaswani, Bengio, Brevdo, Chollet, Gomez, Gouws,
  Jones, Kaiser, Kalchbrenner, Parmar, Sepassi, Shazeer, and
  Uszkoreit}]{vaswani-etal-2018-tensor2tensor}
Ashish Vaswani, Samy Bengio, Eugene Brevdo, Francois Chollet, Aidan Gomez,
  Stephan Gouws, Llion Jones, {\L}ukasz Kaiser, Nal Kalchbrenner, Niki Parmar,
  Ryan Sepassi, Noam Shazeer, and Jakob Uszkoreit. 2018.
\newblock \href {https://www.aclweb.org/anthology/W18-1819} {{T}ensor2{T}ensor
  for neural machine translation}.
\newblock In \emph{Proceedings of the 13th Conference of the Association for
  Machine Translation in the {A}mericas (Volume 1: Research Track)}, pages
  193--199, Boston, MA. Association for Machine Translation in the Americas.

\bibitem[{Vaswani et~al.(2017)Vaswani, Shazeer, Parmar, Uszkoreit, Jones,
  Gomez, Kaiser, and Polosukhin}]{vaswani2017attention}
Ashish Vaswani, Noam Shazeer, Niki Parmar, Jakob Uszkoreit, Llion Jones,
  Aidan~N Gomez, {\L}ukasz Kaiser, and Illia Polosukhin. 2017.
\newblock Attention is all you need.
\newblock In \emph{Advances in neural information processing systems}, pages
  5998--6008.

\bibitem[{Wang et~al.(2020)Wang, Tsvetkov, and
  Neubig}]{wang-etal-2020-balancing}
Xinyi Wang, Yulia Tsvetkov, and Graham Neubig. 2020.
\newblock \href {https://www.aclweb.org/anthology/2020.acl-main.754} {Balancing
  training for multilingual neural machine translation}.
\newblock In \emph{Proceedings of the 58th Annual Meeting of the Association
  for Computational Linguistics}, pages 8526--8537, Online. Association for
  Computational Linguistics.

\bibitem[{Wolf et~al.(2020)Wolf, Debut, Sanh, Chaumond, Delangue, Moi, Cistac,
  Rault, Louf, Funtowicz, Davison, Shleifer, von Platen, Ma, Jernite, Plu, Xu,
  Le~Scao, Gugger, Drame, Lhoest, and Rush}]{wolf-etal-2020-transformers}
Thomas Wolf, Lysandre Debut, Victor Sanh, Julien Chaumond, Clement Delangue,
  Anthony Moi, Pierric Cistac, Tim Rault, Remi Louf, Morgan Funtowicz, Joe
  Davison, Sam Shleifer, Patrick von Platen, Clara Ma, Yacine Jernite, Julien
  Plu, Canwen Xu, Teven Le~Scao, Sylvain Gugger, Mariama Drame, Quentin Lhoest,
  and Alexander Rush. 2020.
\newblock \href {https://doi.org/10.18653/v1/2020.emnlp-demos.6} {Transformers:
  State-of-the-art natural language processing}.
\newblock In \emph{Proceedings of the 2020 Conference on Empirical Methods in
  Natural Language Processing: System Demonstrations}, pages 38--45, Online.
  Association for Computational Linguistics.

\bibitem[{Wu et~al.(2016)Wu, Schuster, Chen, Le, Norouzi, Macherey, Krikun,
  Cao, Gao, Macherey, Klingner, Shah, Johnson, Liu, Kaiser, Gouws, Kato, Kudo,
  Kazawa, Stevens, Kurian, Patil, Wang, Young, Smith, Riesa, Rudnick, Vinyals,
  Corrado, Hughes, and Dean}]{wu-etal-2016-GNMT}
Yonghui Wu, Mike Schuster, Zhifeng Chen, Quoc~V. Le, Mohammad Norouzi, Wolfgang
  Macherey, Maxim Krikun, Yuan Cao, Qin Gao, Klaus Macherey, Jeff Klingner,
  Apurva Shah, Melvin Johnson, Xiaobing Liu, Lukasz Kaiser, Stephan Gouws,
  Yoshikiyo Kato, Taku Kudo, Hideto Kazawa, Keith Stevens, George Kurian,
  Nishant Patil, Wei Wang, Cliff Young, Jason Smith, Jason Riesa, Alex Rudnick,
  Oriol Vinyals, Greg Corrado, Macduff Hughes, and Jeffrey Dean. 2016.
\newblock \href {http://arxiv.org/abs/1609.08144} {Google's neural machine
  translation system: Bridging the gap between human and machine translation}.
\newblock \emph{CoRR}, abs/1609.08144.

\bibitem[{Zaharia et~al.(2016)Zaharia, Xin, Wendell, Das, Armbrust, Dave, Meng,
  Rosen, Venkataraman, Franklin et~al.}]{zaharia2016spark}
Matei Zaharia, Reynold~S Xin, Patrick Wendell, Tathagata Das, Michael Armbrust,
  Ankur Dave, Xiangrui Meng, Josh Rosen, Shivaram Venkataraman, Michael~J
  Franklin, et~al. 2016.
\newblock Apache spark: a unified engine for big data processing.
\newblock \emph{Communications of the ACM}, 59(11):56--65.

\bibitem[{Zhang et~al.(2020)Zhang, Williams, Titov, and
  Sennrich}]{zhang-etal-2020-multiling-nmt}
Biao Zhang, Philip Williams, Ivan Titov, and Rico Sennrich. 2020.
\newblock \href {https://www.aclweb.org/anthology/2020.acl-main.148} {Improving
  massively multilingual neural machine translation and zero-shot translation}.
\newblock In \emph{Proceedings of the 58th Annual Meeting of the Association
  for Computational Linguistics}, pages 1628--1639, Online. Association for
  Computational Linguistics.

\bibitem[{Ziemski et~al.(2016)Ziemski, Junczys-Dowmunt, and
  Pouliquen}]{ziemski-etal-2016-unpc}
Micha{\l} Ziemski, Marcin Junczys-Dowmunt, and Bruno Pouliquen. 2016.
\newblock \href {https://www.aclweb.org/anthology/L16-1561} {The united nations
  parallel corpus v1.0}.
\newblock In \emph{Proceedings of the Tenth International Conference on
  Language Resources and Evaluation ({LREC}'16)}, pages 3530--3534,
  Portoro{\v{z}}, Slovenia. European Language Resources Association (ELRA).

\bibitem[{Zoph et~al.(2016)Zoph, Yuret, May, and
  Knight}]{zoph-etal-2016-transfer}
Barret Zoph, Deniz Yuret, Jonathan May, and Kevin Knight. 2016.
\newblock \href {https://doi.org/10.18653/v1/D16-1163} {Transfer learning for
  low-resource neural machine translation}.
\newblock In \emph{Proceedings of the 2016 Conference on Empirical Methods in
  Natural Language Processing}, pages 1568--1575, Austin, Texas. Association
  for Computational Linguistics.

\end{thebibliography}
\bibliographystyle{acl_natbib}

\end{document}